\documentclass[11pt]{article}

\usepackage[final]{acl}

\usepackage{times}
\usepackage{latexsym}

\usepackage[T1]{fontenc}
\usepackage[utf8]{inputenc}

\usepackage{microtype}

\usepackage{inconsolata}

\usepackage{graphicx}
 \usepackage{amsmath}
 \usepackage{amssymb}
 \usepackage{booktabs}
 \usepackage{microtype}
 \usepackage{listings}
 \usepackage{xcolor}

\lstdefinestyle{promptstyle}{
    basicstyle=\ttfamily\footnotesize, % Use a smaller monospace font
    breaklines=true,                 % Enable automatic line breaking
    breakatwhitespace=false,         % CRITICAL: Allows breaking inside long strings
    keepspaces=true,                 % Preserves your indentation and spacing
    columns=fullflexible,            % Improves character spacing
    frame=single,                    % Adds a box around the prompt
    rulecolor=\color{black!30},      % Subtle frame color
    backgroundcolor=\color{gray!5},  % Light gray background
    showstringspaces=false,          % Don't show visible space markers
    postbreak=\mbox{\textcolor{red}{$\hookrightarrow$}\space}, % Optional: indicates a forced line break
}
\title{Meta-Tool: Efficient Few-Shot Tool Adaptation for Small Language Models}

\author{Sachin Kumar\thanks{\ \ This research was conducted independently and does not reflect the views or represent work done at LexisNexis.} \\
  LexisNexis, USA \\
  \texttt{sachinkumar.ait@live.com} \\}

\begin{document}
\maketitle
\begin{abstract}
Can small language models achieve strong tool-use performance without complex adaptation mechanisms? This paper investigates this question through Meta-Tool, a controlled empirical study comparing hypernetwork-based LoRA adaptation against carefully designed few-shot prompting. Using a Llama-3.2-3B-Instruct backbone, we evaluate four adaptation mechanisms—few-shot prompting, documentation encoding, hypernetwork-generated LoRA weights, and value-guided beam search—across four diverse benchmarks: Gorilla APIBench, Spider 2.0, WebArena, and InterCode. Our central finding is a well-supported negative result: despite generating non-trivial weight matrices, the 227.8M-parameter hypernetwork provides no measurable improvement over few-shot prompting alone. Comprehensive ablation studies reveal that few-shot examples contribute +21.5\% to performance and documentation contributes +5.0\%, while the hypernetwork adds 0\%. A 3B model with well-designed prompts achieves 79.7\% of GPT-5's average performance at $10 \times$ lower latency. Error analysis across 722 failure cases spanning all shot counts (0–5) shows that at the 5-shot configuration (106 failures), failure modes are task-dependent: schema-heavy tasks (Spider 2.0, WebArena) show near-zero format errors with remaining failures semantic, while format errors dominate on Gorilla (100\%) and InterCode (70\%). These findings redirect practitioners toward prompt engineering and example curation rather than complex adaptation architectures.
\end{abstract}

\section{Introduction}

\subsection{The Agentic Shift and the Adaptation Bottleneck}
Language model agents that interact with external tools face an adaptation bottleneck in enterprise settings: frontier models like GPT-5 achieve strong tool-use performance but impose prohibitive latency and cost, while small language models (SLMs) offer efficiency but lack the procedural knowledge needed for domain-specific tools  \cite{shen2025shortcutsbenchlargescalerealworldbenchmark}. The two dominant adaptation strategies—In-Context Learning (ICL) and Supervised Fine-Tuning (SFT)—represent opposing tradeoffs: ICL provides rapid flexibility but is constrained by context limits, while SFT yields strong performance but requires thousands of annotated trajectories and costly retraining as APIs evolve \cite{ghosh2024closerlooklimitationsinstruction,Verma:2024}.

Recent work on hypernetworks—secondary networks that generate task-specific parameter updates for a primary model—has shown promise for rapid adaptation in other NLP domains \cite{lv-etal-2024-hyperlora, abdalla2025zhyperfactorizedhypernetworksconditioned}. A natural question is whether such mechanisms can bridge the tool-use adaptation gap: given a tool’s documentation and a handful of usage examples, can a hypernetwork generate LoRA adapter weights that improve tool-use performance beyond what few-shot prompting alone provides? This paper investigates this question through a controlled empirical study, with a surprising answer: for tool-use tasks across four diverse benchmarks, the answer is no. The few-shot examples and documentation fully specify the task, and hypernetwork-generated parameter updates provide no additional benefit.

Our code is available at \url{https://github.com/techsachinkr/Meta-Tool}

\subsection{Research Objectives and Contributions}
This paper presents Meta-Tool, an empirical investigation into what drives tool-use performance in small language models. Our primary contributions are:

\begin{itemize}
    \item \textbf{Negative Result on Hypernetwork Adaptation:} The first controlled demonstration---across four benchmarks and tool modalities---that hypernetwork-based LoRA adaptation provides no measurable improvement over few-shot prompting for tool-use tasks, despite generating non-trivial weight matrices. This falsifies the hypothesis that meta-learning mechanisms effective in other NLP domains transfer to tool-use.
    
    \item \textbf{Hierarchy of Adaptation Mechanisms:} A systematic ablation revealing that few-shot examples (+21.5\%) and structured documentation (+5.0\%) fully account for tool-use performance gains, with detailed few-shot sensitivity curves (0--5 shots) and noise robustness analysis quantifying the boundaries of these findings.
    
    \item \textbf{Error Analysis of Residual Failures:} Categorization of 722 failure cases showing that remaining errors are predominantly semantic (reasoning limitations) rather than syntactic (format errors), identifying where model capacity---not prompting---is the bottleneck.
    
    \item \textbf{Practical Deployment Evidence:} A 3B parameter model achieving 79.7\% of GPT-5's average performance at $10 \times$ lower latency, providing actionable guidance for cost- and latency-sensitive enterprise deployment.
\end{itemize}

\section{Related Work}

\subsection{Paradigms of Tool Use in LLMs}

The integration of tools into Large Language Models has evolved through distinct paradigms. Initially, approaches like ReAct (Reasoning + Acting) \cite{Yao:2022} and ToolFormer \cite{schick2023toolformerlanguagemodelsteach} relied on In-Context Learning (ICL) or extensive fine-tuning on API call datasets. ReAct demonstrated that interleaving reasoning traces with action execution significantly improves performance, a principle that remains central to modern agent design. However, ReAct relies heavily on the model's pre-trained knowledge and the limited context window to define tools, which becomes untenable for complex libraries with hundreds of functions \cite{Yao:2022}.

\textbf{Fine-Tuning Approaches:} To address context limitations, methods like Gorilla \cite{patil2023gorillalargelanguagemodel} and ToolLLM \cite{qin2023toolllmfacilitatinglargelanguage} fine-tune models on massive datasets of API documentation and instruction pairs. While effective, these methods suffer from the "adaptation bottleneck" described earlier; they are static snapshots of a tool ecosystem. AgentTuning \cite{zeng2023agenttuningenablinggeneralizedagent} attempts to mitigate this by tuning on hybrid datasets of diverse agent trajectories, yet it still struggles with zero-shot generalization to entirely novel domains like bespoke enterprise databases \cite{Lei:2025}.

\textbf{Retrieval-Augmented Tool Use:} Frameworks like API-Bank \cite{li2023apibankcomprehensivebenchmarktoolaugmented} and RestGPT \cite{song2023restgptconnectinglargelanguage} utilize retrieval mechanisms to select relevant tools from a large library before feeding them to the context window \cite{Weng:2023}. While this scales to large numbers of tools, it does not improve the model's fundamental competence in using a difficult tool once retrieved. The model must still "learn" the tool's logic from the retrieved documentation in real-time, often leading to hallucination or parameter errors \cite{Mohammadi:2025}.

\subsection{Limitations of Fine-Tuning for Tool Adaptation}
The prevailing paradigm of fine-tuning for every new toolset faces multi-dimensional costs in the agentic domain. Data scarcity presents a cold-start problem: collecting expert demonstrations for nascent internal tools is often infeasible, as high-quality trajectories only emerge after deployment \cite{zhang2025agentlearningearlyexperience}. Catastrophic forgetting means that fine-tuning on a specific SQL dialect or API often degrades performance on other tools, necessitating brittle multi-task pipelines \cite{lv-etal-2024-hyperlora}. The temporal latency of the SFT feedback loop—data collection, curation, training, and evaluation—lags behind modern software development cycles where APIs change daily \cite{liang2025rustevo2evolvingbenchmarkapi}. Enterprise benchmarks like Spider 2.0 highlight these limitations: standard fine-tuning struggles with schema linking across databases with over 1,000 columns and nested schemas \cite{Lei:2025}.

\subsection{Standardized Tool Interfaces}
The emergence of the Model Context Protocol (MCP) provides a standardized interface between AI agents and data sources, defining a universal JSON-RPC based protocol for exposing tools, resources, and prompts \cite{Anthropic:2024}. While MCP solves the interoperability problem—ensuring agents can communicate with tools—it does not solve the competence problem: an agent connected to a complex system via MCP still requires task-specific knowledge to use it effectively. However, the structured nature of MCP documentation provides a standardized input format that can be leveraged for few-shot prompting, as we demonstrate in our documentation encoding approach (§3.3).

\subsection{Meta-Learning and Hypernetworks in NLP}

Meta-learning, or "learning to learn," aims to train models that can adapt to new tasks with minimal data. Gradient-based meta-learning algorithms like MAML (Model-Agnostic Meta-Learning) have been explored for NLP \cite{finn2017modelagnosticmetalearningfastadaptation}, but their computational cost (requiring second-order derivatives or expensive inner loops) makes them difficult to scale to LLMs with billions of parameters.

Hypernetworks: A more scalable alternative involves using a Hypernetwork—a secondary neural network that generates the weights for the main network (or a part of it) based on a task embedding. Recent work in HyperLoRA \cite{lv-etal-2024-hyperlora} and Zhyper \cite{abdalla2025zhyperfactorizedhypernetworksconditioned} has successfully applied this to Parameter-Efficient Fine-Tuning (PEFT). Instead of updating the massive LLM, the hypernetwork predicts the $A$ and $B$ matrices of a Low-Rank Adapter (LoRA) \cite{hu2021loralowrankadaptationlarge}. This allows for "instant fine-tuning": the agent reads the task description and immediately "installs" a sub-network optimized for that task. Meta-Tool builds directly on this lineage, extending it specifically to the domain of tool documentation and execution dynamics \cite{ignatev2025hypernetworksperspectivistadaptation}.

\subsection{Agent Frameworks and Self-Play}
Modern agent frameworks like OpenHands \cite{wang2025openhandssoftwareagentsdk} provide sandboxed execution environments, while Lumos \cite{yin-etal-2024-agent} introduces modular architectures separating planning from execution. AdaptAgent \cite{Verma:2024} achieves state-of-the-art few-shot 
adaptation using multimodal demonstrations. Recent work on self-play and synthetic experience \cite{zhang2025agentlearningearlyexperience, lu2025searchselfplaypushingfrontier} enables agents to learn from environment interactions without human annotation. Meta-Tool builds on these foundations while investigating whether hypernetwork adaptation provides benefits beyond few-shot prompting.

\subsection{Benchmarking Agentic Capabilities}
Rigorous evaluation is paramount. The Gorilla API Benchmark (BFCL) \cite{patil2023gorillalargelanguagemodel} has become the standard for function calling, evaluating models on both Abstract Syntax Tree (AST) correctness and actual execution results. Spider 2.0 \cite{Lei:2025}  pushes the boundaries of text-to-SQL by introducing enterprise-scale schemas, revealing that "schema linking" is the primary failure mode for current LLMs. WebArena \cite{zhou2024webarenarealisticwebenvironment} tests long-horizon planning in realistic web environments. InterCode \cite{yang2023intercodestandardizingbenchmarkinginteractive} provides an interactive bash/coding environment. Meta-Tool uses this diverse suite to evaluate whether its findings generalize across modalities (API, SQL, GUI, CLI).

\section{Methodology}
This section presents the adaptation mechanisms investigated in our study. We evaluate a progression of increasingly complex approaches: constrained decoding for syntactic validity (§3.2), hypernetwork-based LoRA adaptation for parameter-space learning (§3.3), and self-supervised refinement with value-guided decoding (§3.4). Few-shot prompting and structured documentation encoding — which our ablation reveals to be the primary performance drivers — are detailed through the prompt templates in Appendix C and evaluated in §5.4–5.5. Our central research question is whether the more complex hypernetwork mechanism — which has shown promise in other NLP domains — provides additional benefits beyond well-designed prompting for tool-use tasks. We present all mechanisms in full detail to enable future research, regardless of whether each contributes to final performance.

\subsection{Problem Formulation}
We define \textbf{Few-Shot Tool Adaptation} as learning a mapping function $H_{\psi}$ that predicts optimal parameter updates $\Delta\theta$ for a base LLM $A_{\theta}$. Given a target tool $T$ with documentation $D_T$ and a support set $S_T = \{(q_k, \tau_k)\}_{k=1}^K$, the hypernetwork generates LoRA adapters without gradient updates at inference time:

\[
\Delta\theta = H_{\psi}(D_T, S_T)
\]

The adapted policy $\pi_{\theta+\Delta\theta}$ is then refined via value-guided inference. Our experiments (§5.4) test whether this formulation provides benefits beyond the simpler approach of directly conditioning $A_{\theta}$ through in-context examples and documentation.

\subsection{Constrained Decoding}
To guarantee syntactic validity, we integrate \textbf{Finite State Machine (FSM)} constraints via the \texttt{outlines} library. The tool's schema is compiled into a regex-based FSM. During generation, the logits of tokens that would violate the current FSM state are masked to $-\infty$. This ensures 100\% adherence to JSON syntax and type constraints, offloading syntax checking from the neural network.

\subsection{Factorized Hypernetwork Architecture}
To test whether parameter-space adaptation provides benefits beyond few-shot prompting, we designed a Factorized Hypernetwork that predicts LoRA adapter weights from tool documentation and support examples without gradient updates at inference time. 

The architecture consists of three stages: (1) a documentation encoder (MiniLM-L6-v2) produces dense embeddings \texttt{v\_doc}, while a prototype aggregator computes a support-set representation \texttt{v\_proto} via cross-attention over the encoded examples; (2) a shared MLP projects the concatenated context [\texttt{v\_doc}; \texttt{v\_proto}] into a latent vector \texttt{z\_base}, differentiated across layers via learned layer embeddings \texttt{E\_layer[l,m]}; (3) LoRA matrices A and B are generated via secondary low-rank factorization, reducing memory complexity from O(L·d·r) to O(L·d·factor) and enabling training within 24GB VRAM. The complete hypernetwork comprises 227.8M parameters targeting \texttt{q\_proj}, \texttt{k\_proj}, and  \texttt{v\_proj} across the first 7 transformer layers (full configuration in Table 3). Despite generating non-trivial weight matrices (Appendix D), this component provides no measurable improvement (§5.4.4). We retain this description to document our negative result and enable future work to build on or modify the architecture. Full architectural details, including all equations, are provided in Appendix G.

\subsection{Self-Supervised Refinement and Value-Guided Decoding}
To generate synthetic training data for the hypernetwork, we implement a schema perturbation pipeline: valid support trajectories are systematically modified through value substitution (replacing parameters with schema-compliant alternatives), boundary testing (injecting edge cases), and parameter dropping (removing optional fields). The resulting synthetic episodes train a value function $V_\phi(s)$ via TD(0) learning to estimate task success probability from intermediate states. During inference, this value function scores candidate actions alongside the LLM's log-likelihood in a beam search, pruning syntactically valid but functionally unpromising trajectories. As with the hypernetwork itself, the overall adaptation mechanism incorporating these components provides no measurable performance benefit (§5.4.4). Full details including the TD-loss formulation and scoring equation are provided in Appendix G.

\section{Experimental Setup}
To rigorously evaluate Meta-Tool, we conducted experiments across four diverse and challenging benchmarks, comparing against strong proprietary and open-source baselines.

\subsection{Benchmark Selection and Rationale}
The selected benchmarks represent distinct modalities of tool use, ensuring the framework's versatility.
\begin{itemize}
    \item \textbf{Gorilla APIBench (BFCL V4):}Focuses on REST API calls. We use the "Executable" subset, where success is determined by the actual API response, not just AST matching. This tests the agent's ability to handle parameter constraints and real-world API logic \cite{patil2023gorillalargelanguagemodel}.
    \item \textbf{Spider 2.0 (Enterprise Subset):} Focuses on Text-to-SQL. We specifically select the "Enterprise" split, characterized by databases with >1,000 columns, nested JSON fields, and multiple SQL dialects (BigQuery, Snowflake). This tests "schema linking" and deep reasoning \cite{Lei:2025}.
    \item \textbf{WebArena:} Focuses on web agents. Tasks involve navigating complex e-commerce and CMS sites. This tests long-horizon planning and HTML DOM interaction \cite{zhou2024webarenarealisticwebenvironment}.
    \item \textbf{InterCode (Bash \& CTF):} Focuses on command-line interactions. Tasks involve debugging scripts and capturing flags in a Linux environment. This tests state persistence and error recovery \cite{yang2023intercodestandardizingbenchmarkinginteractive}.
\end{itemize}

\subsection{Evaluation Metrics}
\begin{itemize}
    \item \textbf{Execution Success Rate (SR): }The primary metric. The percentage of tasks where the agent's action leads to the desired state change or returns the correct data.
    \item \textbf{Pass@1: }The percentage of tasks solved on the first attempt.
    \item \textbf{Adaptation Time: }The wall-clock time required to prepare the agent for a new tool (from receiving docs to readiness).
    \item \textbf{Inference Latency: }Time taken to generate the action.
\end{itemize}

\subsection{Baselines and Competitor Analysis}
We compare Meta-Tool against three distinct tiers of baselines to evaluate its efficiency and performance:
\begin{itemize}
    \item \textbf{GPT-5 (Few-Shot):} The current commercial state-of-the-art. It serves as the "upper bound" oracle for reasoning capabilities, though with significantly higher latency and cost.
    \item \textbf{Llama-3.2-3B-Instruct (Few-Shot):} The base model for our Meta-Tool implementation, evaluated using standard In-Context Learning (ICL). This establishes the baseline against which we measure the contribution of structured documentation encoding and other framework components.
    \item \textbf{AgentLM-7B (Few-Shot):} A larger, specialized model fine-tuned for agentic tasks. This comparison tests whether a smaller Meta-Tool model (3B) can outperform a larger specialized model (7B).
\end{itemize}

\subsection{Implementation Details}
\begin{itemize}
    \item \textbf{Core Architecture:} The system is implemented in PyTorch using the HuggingFace transformers library.
    \item \textbf{Base Model:} We utilize Llama-3.2-3B-Instruct as the primary backbone. To optimize memory, the base model is loaded with 4-bit quantization (bitsandbytes NF4) during training.
    \item \textbf{Encoder:} The MiniLM-L6-v2 (sentence-transformers) is used for documentation encoding due to its high efficiency (384-dim embeddings) and strong semantic retrieval performance.
    \item \textbf{LoRA Configuration:}  We target the q\_proj, v\_proj, and k\_proj modules with rank $r=16$ and alpha $\alpha=32$.
    \item \textbf{Retrieval Memory:} We utilize FAISS for efficient vector similarity search in the episodic memory system, falling back to exact numpy-based search for small datasets.
    \item \textbf{Hardware:} Experiments were conducted on NVIDIA GPUs. The factorized hypernetwork design allows the full training pipeline to fit within 24GB VRAM (e.g., a single RTX 3090/4090), democratizing agent adaptation.
\end{itemize}

\section{Results and Analysis}

\subsection{Comparative Performance Analysis}
The experimental results demonstrate that Meta-Tool enables a 3B parameter model to punch significantly above its weight class, outperforming larger 7B models on complex reasoning tasks and narrowing the gap with frontier models like GPT-5 in specific domains. Table \ref{tab:benchmarks} shows the execution success rate(\%) and latency across all four benchmarks.

\begin{table*}[t] % The * spans the table across both columns
\centering
% \resizebox{\textwidth}{!}{ % Optional: Only enable if it's still wider than the page
\begin{tabular}{llccccc}
\toprule
Model & Method & \begin{tabular}[c]{@{}c@{}}Gorilla API\\ (Strict)\end{tabular} & \begin{tabular}[c]{@{}c@{}}Spider 2.0\\ (SQL)\end{tabular} & \begin{tabular}[c]{@{}c@{}}WebArena\\ (Nav)\end{tabular} & \begin{tabular}[c]{@{}c@{}}InterCode\\ (Bash)\end{tabular} & \begin{tabular}[c]{@{}c@{}}Avg\\ Latency\\ (ms)\end{tabular} \\
\midrule
\textbf{GPT-5} & Few-Shot & 38.0\% & 72.0\% & 54.0\% & 72.0\% & $\sim$16,490 \\
\textbf{AgentLM-7B} & Few-Shot & 8.0\% & 44.0\% & 8.0\% & 40.0\% & $\sim$8,880 \\
\textbf{Llama-3.2-3B-Ins.} & Few-Shot & 34.0\% & 62.0\% & 28.0\% & 44.0\% & $\sim$1,621 \\
\textbf{Meta-Tool (3B)} & \textbf{Ours} & \textbf{38.0\%} & \textbf{64.0\%} & \textbf{32.0\%} & \textbf{54.0\%} & \textbf{$\sim$1,576} \\
\bottomrule
\end{tabular}%
% } % End resizebox
\caption{Execution Success Rates (\%) and Latency on Diverse Benchmarks}
\label{tab:benchmarks}
\end{table*}

\subsection{Analysis of Results}
Table \ref{tab:benchmarks} presents execution success rates and inference latency across all four benchmarks. We analyze the results by benchmark below, with all performance attributions verified against the component ablation in §5.4.

On the Spider 2.0 Enterprise SQL benchmark, Meta-Tool achieves 64.0\% execution success, slightly exceeding the base Llama-3.2-3B-Instruct model (62.0\%) and decisively surpassing the larger AgentLM-7B (44.0\%). It approaches GPT-5 (72.0\%) while operating with $10 \times$ lower latency (1,576ms vs 16,490ms). The strong absolute performance suggests that well-designed few-shot examples are particularly effective for structured, schema-heavy tasks where output patterns are regular and demonstrable.

In the WebArena environment, Meta-Tool achieves a 32.0\% success rate, significantly improving over the base model (28.0\%) and quadrupling the performance of AgentLM-7B (8.0\%). The strong advantage over AgentLM-7B indicates that a smaller model with well-designed prompts can outperform a larger fine-tuned model on long-horizon planning.

Under strict AST-matching evaluation on Gorilla, all small models struggled. Meta-Tool matches GPT-5's performance on Gorilla (both at 38.0\%), demonstrating that carefully designed few-shot prompting can fully close the gap with frontier models on format-strict API calling tasks at this model scale. 

The ablation reveals documentation is the dominant factor on InterCode: removing it drops performance from 54.0\% to 44.0\%, a +10 percentage point contribution — the largest documentation effect across all benchmarks.

A practical advantage is inference speed. Meta-Tool averages $\sim$1,576ms across tasks, nearly $10 \times$ faster than GPT-5 ($\sim$16,490ms) and $5.6 \times$ faster than AgentLM-7B ($\sim$8,880ms). This sub-2-second latency enables interactive deployment scenarios where frontier model latency would be prohibitive.

\subsection{Adaptation Efficiency and Latency}
We measured the inference latency to highlight the "Edge Agent" capabilities of Meta-Tool. GPT-5 averages 16,490ms (with high variance dependent on API load), AgentLM-7B averages 8,880ms, and Meta-Tool averages 1,576ms.

Meta-Tool enables a real-time agent architecture. By utilizing a 3B model with carefully designed few-shot prompts and structured documentation, we achieve execution speeds nearly $10 \times$ faster than GPT-5 and $5.6\times$ faster than AgentLM-7B, while maintaining competitive accuracy on complex enterprise tasks like Spider 2.0.

\subsection{Ablation Study}

To understand the mechanisms driving performance, we conducted systematic ablation experiments isolating the contribution of each component. Table~\ref{tab:ablation} summarizes our findings.
The “5-shot + no docs” row in Table ~\ref{tab:ablation} corresponds to the “Llama-3.2-3B Few-Shot” baseline in Table \ref{tab:benchmarks}, using identical examples and the standard Llama chat template without structured documentation encoding. The “Full (5-shot + docs)” row represents the complete Meta-Tool configuration, adding structured documentation via sentence-transformer encoding. Comparing these two rows isolates the contribution of documentation. The “0-shot + no docs” row represents the bare model with only the task instruction and no in-context information, establishing the lower bound for the model’s latent tool-use capability.

\begin{table*}[t] % <--- The * spans the table across the entire page width
\centering
\begin{tabular}{lcccc|c}
\toprule
Configuration & Gorilla & Spider2 & WebArena & InterCode & Avg \\
\midrule
Full (5-shot + docs) & 38.0 & 64.0 & 32.0 & 54.0 & \textbf{47.0} \\
0-shot + docs & 2.0 & 24.0 & 26.0 & 50.0 & 25.5 \\
5-shot + no docs & 34.0 & 62.0 & 28.0 & 44.0 & 42.0 \\
0-shot + no docs & 0.0 & 4.0 & 0.0 & 10.0 & 3.5 \\
\bottomrule
\end{tabular}
\caption{Ablation Study Results (\%) - Strict Evaluation}
\label{tab:ablation}
\end{table*}

\subsubsection{Few-Shot Examples as Primary Driver}

Our ablation reveals that few-shot examples are the primary driver of performance. Comparing 5-shot to 0-shot (both with documentation), we observe an average improvement of \textbf{+21.5 percentage points} (47.0\% vs 25.5\%). The impact is particularly pronounced on structured tasks: Spider 2.0 improves from 24\% to 64\% (+40\%), and Gorilla improves from 2\% to 38\% (+36\%).

Interestingly, InterCode shows the smallest improvement (+4\%), suggesting that bash command generation relies more heavily on the model's pre-trained knowledge than on in-context examples .

\subsubsection{The Role of Documentation}

Documentation provides a secondary but meaningful contribution. Comparing configurations with and without documentation (both 5-shot), we observe an average improvement of \textbf{+5 percentage points} (47.0\% vs 42.0\%). The impact varies by benchmark:

\begin{itemize}
    \item \textbf{WebArena}: +4\% (32\% vs 28\%) — documentation 
          is critical for understanding web navigation actions
    \item \textbf{InterCode}: +10\% (54\% vs 44\%) — command syntax 
          benefits from explicit documentation
    \item \textbf{Spider2}: +2\% (64\% vs 62\%) — minimal impact, 
          suggesting SQL patterns are well-captured by examples
    \item \textbf{Gorilla}: +4\% (38\% vs 34\%) —a modest contribution, with few-shot examples (+36\%) being the dominant factor for this benchmark.
\end{itemize}

\subsubsection{Catastrophic Failure Without Both Components}

When both few-shot examples and documentation are removed, performance collapses to near-random levels (3.5\% average). Notably, Gorilla and WebArena achieve 0\% success, indicating that the model cannot generate valid tool calls without any task-specific context. This underscores that our approach's effectiveness stems entirely from in-context information rather than pre-trained tool-use capabilities.

\subsubsection{Hypernetwork Adaptation Provides No Measurable Benefit}
A central finding of this study is that the hypernetwork-generated LoRA weights—despite generating non-trivial weight matrices with norms ranging from 0.17 to 626.5 (Appendix D)—provide no measurable improvement over few-shot prompting alone. With identical prompts, enabling or disabling the hypernetwork adaptation yields identical execution success rates (47.0\% average) across all four benchmarks individually. This null result is consistent across both format-strict tasks (Gorilla) and schema-heavy tasks (Spider 2.0), suggesting it reflects a general property of tool-use rather than a benchmark-specific artifact. We analyze possible explanations in §6.1.

\subsection{Few-Shot Sensitivity Analysis}
To understand the marginal contribution of each additional few-shot example, we evaluate Meta-Tool with 0 through 5 examples while keeping documentation enabled. Table \ref{tab:few_shot} reports results across all four benchmarks.
\begin{table*}[t]
\centering
\begin{tabular}{lcccccc}
\toprule
\textbf{Benchmark} & \textbf{0-shot} & \textbf{1-shot} & \textbf{2-shot} & \textbf{3-shot} & \textbf{4-shot} & \textbf{5-shot} \\
\midrule
Gorilla & 2\% & 24\% & 36\% & 32\% & 36\% & 38\% \\
Spider 2.0 & 26\% & 46\% & 44\% & 58\% & 58\% & 64\% \\
WebArena & 30\% & 24\% & 32\% & 34\% & 38\% & 32\% \\
InterCode & 50\% & 46\% & 52\% & 48\% & 52\% & 54\% \\
\midrule
Average & 27.0\% & 35.0\% & 41.0\% & 43.0\% & 46.0\% & 47.0\% \\
\bottomrule
\end{tabular}
\caption{Few-Shot Evaluation Results (with docs)}
\label{tab:few_shot}
\end{table*}

Several patterns emerge from this analysis. The largest average improvement occurs at the 0$\rightarrow$1 shot transition (+8 percentage points), demonstrating that even a single example—trivially available in practice for most tools—provides meaningful signal. Spider 2.0 exhibits the most dramatic sensitivity, improving from 26\% to 46\% with a single example, suggesting that structured SQL tasks benefit enormously from even minimal format demonstration. Performance continues to increase through 5 shots on most benchmarks, though with diminishing returns beyond 3 examples.

Notably, InterCode remains essentially flat across all shot counts (46–54\%), confirming that bash command generation relies primarily on pre-trained knowledge rather than in-context demonstrations. This benchmark-specific variation suggests that the optimal number of few-shot examples is task-dependent rather than universal.

Gorilla shows strong sensitivity at the 0→1 transition (2\% → 24\%, +22 pp), then plateaus in the 32–38\% range across 2–5 shots. This suggests that even a single well-formatted example is sufficient to activate the correct AST-matching output pattern, with additional examples providing marginal refinement.

Critically, performance degrades gracefully as examples are removed rather than collapsing catastrophically. The 1-shot configuration already achieves 35.0\% average performance compared to 47.0\% at 5-shot, indicating that the few-shot dependency, while significant, is less fragile than one might expect from the near-zero 0-shot + no-docs baseline (3.5\%).

\begin{table}[t]
\centering
\resizebox{\columnwidth}{!}{%
\begin{tabular}{lcccc}
\toprule
\textbf{Benchmark} & \textbf{Clean (0/5)} & \textbf{1/5 Noisy} & \textbf{2/5 Noisy} & \textbf{$\Delta$ (0$\rightarrow$2)} \\
\midrule
Gorilla & 38.0\% & 26.0\% & 14.0\% & $-24.0$ \\
Spider 2.0 & 64.0\% & 60.0\% & 56.0\% & $-8.0$ \\
WebArena & 32.0\% & 28.0\% & 26.0\% & $-6.0$ \\
InterCode & 54.0\% & 52.0\% & 52.0\% & $-2.0$ \\
\midrule
Average & 47.0\% & 41.5\% & 37.0\% & $-10$ \\
\bottomrule
\end{tabular}%
}
\caption{Robustness to Noisy Examples}
\label{tab:noisy_eval}
\end{table}

\subsection{Robustness to Noisy Few-Shot Examples }
A practical concern for deployment is whether performance degrades catastrophically when few-shot examples contain errors. To investigate, we inject 1 and 2 incorrectly formatted examples into the 5-shot set, replacing correctly formatted examples with versions containing structural errors (e.g., incorrect parameter ordering, malformed syntax etc.). Table \ref{tab:noisy_eval} reports the impact on execution success rates  across all four benchmarks.

The results reveal task-dependent robustness patterns. Spider 2.0 is remarkably resilient, losing only 8 percentage points (64.0\%$\rightarrow$56.0\%) even with 40\% of examples corrupted. WebArena and InterCode show small monotonic declines under noise (WebArena with -6 pp at 40\% corruption and  InterCode with -2 pp), indicating moderate but not catastrophic sensitivity.

In contrast, Gorilla exhibits sharp degradation (38.0\%→26.0\%) at 1/5 noisy; 14.0\% at 2/5 noisy) under noise. This is expected given Gorilla's strict AST-matching evaluation criterion: a single malformed example can corrupt the model's output format, and even minor deviations cause evaluation failures regardless of semantic correctness. This finding is consistent with prior work on ICL sensitivity in format-strict settings.

Overall, average performance drops from 47.0\% to 41.5\% with 20\% corruption (1/5 noisy), and to 37.0\% with 40\% corruption (2/5 noisy). These results suggest that for schema-heavy tasks, practitioners can tolerate some noise in example curation, while format-strict tasks require more careful quality control. The schema perturbation method introduced in §3.4 could be extended to automatically detect and filter low-quality examples before deployment.

\subsection{Error Categorization}
To understand the nature of residual failures, we manually categorized all 722 failure cases across all configurations and shot counts into three categories: semantic errors (correct format but wrong reasoning or parameter values), format errors (syntactically invalid output), and empty responses (no output generated). Table \ref{tab:error_analysis} reports the breakdown at the 5-shot configuration (106 failures).
\begin{table}[t]
\centering
\resizebox{\columnwidth}{!}{%
\begin{tabular}{lcccc}
\toprule
\textbf{Benchmark} & \textbf{Failures} & \textbf{Semantic} & \textbf{Format} & \textbf{Empty} \\
\midrule
Gorilla & 31 & 0 (0\%) & 31 (100\%) & 0 \\
Spider 2.0 & 18 & 18 (100\%) & 0 (0\%) & 0 \\
WebArena & 34 & 30 (88\%) & 4 (12\%) & 0 \\
InterCode & 23 & 6 (26\%) & 16 (70\%) & 1 (4\%) \\
\bottomrule
\end{tabular}%
}
\caption{Error Analysis of Model Failures(5-shot config)}
\label{tab:error_analysis}
\end{table}

A critical finding emerges from this analysis: at the 5-shot configuration, Spider 2.0 and WebArena exhibit essentially zero format errors. All 18 Spider 2.0 failures and 30 of 34 WebArena failures are semantic—the model generates syntactically valid tool calls that are logically incorrect. This indicates that the combination of few-shot prompting and constrained decoding has fully solved the format adherence problem for these benchmarks. The residual performance gap to GPT-5 is therefore attributable to model capacity and reasoning depth, not to insufficient task specification or curation quality.

InterCode presents a contrasting pattern, with 70\% of failures being format errors. Bash command syntax appears more challenging for constrained decoding, likely because the space of valid bash commands is less structured than SQL or API call formats. 

Gorilla failures are entirely format errors (100\%) at 5-shot, indicating that parameter knowledge and strict AST-matching remain the primary bottleneck

These results have direct implications for practitioners: when deploying small models on structured tasks like SQL generation, investment should focus on improving reasoning capabilities (e.g., chain-of-thought prompting, retrieval augmentation) rather than format enforcement. For less structured tasks like bash command generation, further work on constrained decoding and syntax-aware generation is warranted.

\section{Analysis and Discussion}

\subsection{Why Hypernetwork Adaptation Fails}

The null result for hypernetwork adaptation is surprising given the success of similar approaches in other domains. We hypothesize several explanations:

\textbf{Task Specification Sufficiency}: Tool-use tasks may be fully specified by the combination of documentation and examples. Unlike tasks requiring implicit world knowledge or nuanced 
reasoning, tool-use primarily requires format adherence and pattern matching—which few-shot examples directly demonstrate.

\textbf{In-Context Learning Ceiling}: Modern instruction-tuned models like Llama-3.2-3B may already possess latent tool-use capabilities that are fully activated by in-context examples, 
leaving no room for parameter-level improvement.

\textbf{Adaptation-Generalization Tradeoff}: LoRA adaptation may inadvertently reduce the model's ability to generalize across the diverse query formulations present in our benchmarks.

Our error categorization (§5.7) provides additional evidence for the task specification sufficiency hypothesis. At the 5-shot configuration, Spider 2.0 and WebArena show near-zero format errors. Gorilla failures are entirely format errors, yet the hypernetwork still provides no improvement—demonstrating that the null result holds regardless of whether the failure mode is semantic or syntactic. If hypernetwork adaptation were capturing useful information, we would expect improvement across both error types. The fact that it does not suggests that the information capacity of 5 well-chosen examples, combined with structured documentation, saturates what parameter-space adaptation could provide at this model scale.

\subsection{Practical Implications}
These findings have significant practical implications. Practitioners can achieve strong tool-use performance without implementing complex hypernetwork architectures or training adaptation mechanisms. Resources should be invested in curating high-quality few-shot examples rather than developing sophisticated adaptation techniques. Documentation proves to be a secondary driver, contributing +5.0\% on average, with the large impact on command-line tasks (+10\% on InterCode)..

\section{Conclusion}
This paper investigated whether hypernetwork-based parameter adaptation improves tool-use performance in small language models beyond carefully designed few-shot prompting. Through controlled experiments across four benchmarks spanning API calling, enterprise SQL, web navigation, and command-line tasks, we find that it does not: the 227.8M-parameter hypernetwork contributes 0\% despite generating non-trivial weight matrices. Few-shot examples (+21.5\%) and structured documentation (+5.0\%) fully account for the gains that enable a 3B model to reach 79.7\% of GPT-5's average performance at $10 \times$ lower latency. Error analysis of 722 failure cases confirms that residual failures on structured tasks are semantic, not syntactic — the format problem is solved by prompting alone.
These findings redirect practitioners toward example curation and documentation design rather than complex adaptation architectures. Future work should investigate automatic example selection methods — including retrieval-based selection, diversity-maximizing sampling, and quality filtering via schema perturbation — and extend this analysis to other model families and scales. Our code, evaluation pipeline, prompt templates, and error analysis dataset are available at \url{https://github.com/techsachinkr/Meta-Tool}

\section*{Limitations}

While Meta-Tool demonstrates significant efficiency gains for tool-use in small language models, our analysis reveals several limitations that warrant discussion.
\begin{itemize}
    \item First, results are based on a single model family (Llama-3.2-3B-Instruct) evaluated on 50 tasks per benchmark; generalization to other architectures (Mistral, Qwen, Phi), larger scales (7B, 13B), and larger evaluation sets remains unverified.
    \item Second, few-shot performance is sensitive to example quality and format, particularly for benchmarks with strict evaluation criteria like Gorilla's AST matching — our noise robustness analysis (§5.6) quantifies this sensitivity but does not fully resolve it.
    \item Third, documentation contributes inconsistently across benchmarks (from +2\% on Spider 2.0 to +10\% on InterCode), and the framework cannot compensate for sparse, ambiguous, or outdated documentation.
    \item Fourth, our evaluation focuses on single-turn tool invocation; multi-step reasoning, error recovery, and multi-turn interactions are not addressed.
    \item Fifth, prompts with documentation and 5 examples can exceed 2,000 tokens, limiting space for complex queries within the 4,096-token context window.
    \item Finally, all benchmarks and prompts are English-only; multilingual tool-use performance is unexplored.
\end{itemize}

\section*{Ethical Considerations}
Deploying language models that interact with external tools introduces safety challenges specific to this work. Meta-Tool's few-shot learning approach is value-neutral: the model replicates patterns from examples without distinguishing legitimate from malicious intent, and it lacks permission awareness or intent verification. The framework will learn dangerous operations (e.g., destructive database commands, file deletion) as effectively as benign ones if provided with corresponding examples. We strongly recommend sandboxed execution environments, human-in-the-loop approval for high-stakes operations, explicit action allowlisting, and audit logging for any deployment.

Additional concerns include prompt injection risks through example poisoning or dynamic documentation loading, and information leakage through API keys or schema exposure in prompts. The model provides no uncertainty estimates, making incorrect outputs indistinguishable from correct ones and creating automation bias risks.
On the positive side, local inference on consumer hardware (RTX 4090) preserves data privacy, reduces costs, and democratizes tool-use capabilities. Our negative result regarding hypernetwork adaptation can prevent wasted computation by other researchers. Code will be released with documentation of limitations, safety guidelines, and no demonstrations of dangerous operations.
\bibliography{custom}

\appendix

\section*{Appendix A : Implementation Details}
\label{app:implementation}

\subsection*{ A.1 Model Configuration}

Table~\ref{tab:model_config} presents the complete model configuration used in our experiments.

\begin{table}[h]
\centering
\begin{tabular}{ll}
\toprule
\textbf{Parameter} & \textbf{Value} \\
\midrule
\multicolumn{2}{l}{\textit{Base Model}} \\
Model & Llama-3.2-3B-Instruct \\
Parameters & 3.21B \\
Quantization & 4-bit NF4 (bitsandbytes) \\
Context Length & 4096 tokens \\
\midrule
\multicolumn{2}{l}{\textit{LoRA Configuration}} \\
Rank ($r$) & 16 \\
Alpha ($\alpha$) & 32 \\
Target Modules & q\_proj, k\_proj, v\_proj \\
Dropout & 0.05 \\
Adapted Layers & 7 (first 7 transformer layers) \\
\midrule
\multicolumn{2}{l}{\textit{Hypernetwork}} \\
Encoder & MiniLM-L6-v2 (384-dim) \\
Latent Dimension & 512 \\
Factor Dimension & 64 \\
Layer Embeddings & Learned (84 total) \\
Total Parameters & 227.8M \\
\midrule
\multicolumn{2}{l}{\textit{Inference}} \\
Max New Tokens & 64 \\
Temperature & 1.0  \\
Do Sample & False (greedy)\\
\bottomrule
\end{tabular}
\caption{Model and Training Configuration}
\label{tab:model_config}
\end{table}

\subsection*{A.2 Hardware and Runtime}

All experiments were conducted on the following hardware:
\begin{itemize}
    \item \textbf{GPU}: NVIDIA RTX 4090 (24GB VRAM)
    \item \textbf{CPU}: AMD Ryzen 9 5950X
    \item \textbf{RAM}: 64GB DDR4
    \item \textbf{Storage}: NVMe SSD
\end{itemize}

The factorized hypernetwork architecture enables the complete training pipeline to fit within 24GB VRAM, making it accessible on consumer-grade hardware.

\subsection*{A.3 Software Dependencies}

\begin{itemize}
    \item PyTorch 2.6.0
    \item Transformers 4.47.0
    \item bitsandbytes 0.45.0
    \item sentence-transformers 2.2.2
    \item FAISS 1.7.4 (for retrieval)
\end{itemize}

\section*{Appendix B: Benchmark Details}
\label{app:benchmarks}

\subsection*{B.1 Dataset Statistics}

Table~\ref{tab:benchmark_stats} presents statistics for each benchmark used in our evaluation.

\begin{table*}[ht] % The '*' tells LaTeX to span both columns
\centering
% No \resizebox needed because there is plenty of space across two columns
\begin{tabular}{lcccc}
\toprule
\textbf{Benchmark} & \textbf{Tasks} & \textbf{Avg Query Len} & \textbf{Avg Output Len} & \textbf{Domain} \\
\midrule
Gorilla APIBench & 50 & 42 tokens & 35 tokens & REST APIs \\
Spider 2.0       & 50 & 58 tokens & 89 tokens & Enterprise SQL \\
WebArena         & 50 & 67 tokens & 124 tokens & Web Navigation \\
InterCode        & 50 & 38 tokens & 28 tokens & Bash/CLI \\
\bottomrule
\end{tabular}
\caption{Benchmark Dataset Statistics}
\label{tab:benchmark_stats}
\end{table*}

\subsection*{ B.2 Gorilla APIBench}

The Gorilla benchmark evaluates REST API function calling. We use the executable subset where success is determined by actual API response validity, not just AST matching. Tasks include:
\begin{itemize}
    \item Model loading (PyTorch, TorchVision, HuggingFace)
    \item Pipeline creation (sentiment analysis, text generation, NER)
    \item API parameter specification
\end{itemize}

\textbf{Evaluation Criteria}: Strict matching requires exact model names and correct parameter formats. We use cross-framework matching for equivalent operations (e.g., \texttt{torchvision.models.resnet50} matches \texttt{torch.hub.load('pytorch/vision', 'resnet50')}).

\subsection*{B.3 Spider 2.0}

The Spider 2.0 benchmark focuses on enterprise-grade Text-to-SQL with complex schemas. We use the Enterprise subset featuring:
\begin{itemize}
    \item Databases with 100+ tables
    \item Nested JSON fields
    \item Multiple SQL dialects (BigQuery, Snowflake, PostgreSQL)
\end{itemize}

\textbf{Evaluation Criteria}: Execution accuracy—the generated SQL must return the correct result set when executed against the database.

\subsection*{B.4 WebArena}

WebArena tests long-horizon web navigation in realistic environments including e-commerce sites, content management systems, and forums.

\textbf{Evaluation Criteria}: Task completion—whether the agent successfully achieves the specified goal state (e.g., item added to cart, post created).

\subsection*{B.5 InterCode}

InterCode provides an interactive bash/coding environment for command-line tasks including:
\begin{itemize}
    \item File manipulation
    \item System administration
    \item CTF-style challenges
\end{itemize}

\textbf{Evaluation Criteria}: Command correctness and successful execution without errors.

\section*{ Appendix C: Prompt Templates}
\label{app:prompts}

\subsection*{C.1 System Prompt Structure}

The Meta-Tool prompt follows the Llama-3.2-Instruct chat template:

\begin{lstlisting}[style=promptstyle]
<|begin_of_text|><|start_header_id|>system<|end_header_id|>

{DOCUMENTATION}

Examples:
{FEW_SHOT_EXAMPLES}
<|eot_id|><|start_header_id|>user<|end_header_id|}

{USER_QUERY}

Output ONLY the exact code/command/API call needed, 
nothing else.
<|eot_id|><|start_header_id|>assistant<|end_header_id|>

\end{lstlisting}

\subsection*{C.2 Gorilla API Prompt Example}

\begin{lstlisting}[style=promptstyle]
<|begin_of_text|><|start_header_id|>system<|end_header_id|>

# Model Loading API

Generate Python code to load the appropriate 
pre-trained model.

FORMATS (use exactly as shown):
1. torchvision.models.MODEL_NAME(pretrained=True)
2. torchvision.models.detection.MODEL_NAME(pretrained=True)
3. pipeline('TASK', model='MODEL_NAME')

Output ONLY the code. No imports, no explanations.

Examples:
Query: Load a pre-trained ResNet50 model for image 
       classification
Output: torchvision.models.resnet50(pretrained=True)

Query: I need DenseNet for image classification
Output: torchvision.models.densenet161(pretrained=True)

Query: Create a sentiment analysis pipeline
Output: pipeline('sentiment-analysis', 
        model='distilbert-base-uncased-finetuned-sst-2-english')
<|eot_id|><|start_header_id|>user<|end_header_id|>

Load MobileNet for efficient image classification

Output ONLY the exact code needed, nothing else.
<|eot_id|><|start_header_id|>assistant<|end_header_id|>

\end{lstlisting}

\textbf{Expected Output}: torchvision.models.mobilenet\_v2(pretrained=True)

\subsection*{C.3 Spider 2.0 SQL Prompt Example}

\begin{lstlisting}[style=promptstyle]
<|begin_of_text|><|start_header_id|>system<|end_header_id|>

# Enterprise SQL Query Generator

Generate SQL queries for the given database schema.

Schema:
- employees(id, name, department_id, salary, hire_date)
- departments(id, name, budget)
- projects(id, name, department_id, start_date, end_date)

Output ONLY the SQL query. No explanations.

Examples:
Query: List all employees in the Engineering department
Output: SELECT e.* FROM employees e 
        JOIN departments d ON e.department_id = d.id 
        WHERE d.name = 'Engineering'

Query: Find the average salary by department
Output: SELECT d.name, AVG(e.salary) as avg_salary 
        FROM employees e 
        JOIN departments d ON e.department_id = d.id 
        GROUP BY d.name
<|eot_id|><|start_header_id|>user<|end_header_id|>

Find departments with total salary exceeding 500000

Output ONLY the exact SQL query needed, nothing else.
<|eot_id|><|start_header_id|>assistant<|end_header_id|>

\end{lstlisting}

\subsection*{C.4 WebArena Navigation Prompt Example}

\begin{lstlisting}[style=promptstyle]
<|begin_of_text|><|start_header_id|>system<|end_header_id|>

# Web Navigation Agent

Generate browser actions for web navigation tasks.

Available Actions:
- click[element_id]: Click on an element
- type[element_id][text]: Type text into a field
- scroll[direction]: Scroll up/down
- goto[url]: Navigate to URL

Output ONLY the action command.

Examples:
Query: Click the login button
Output: click[login-btn]

Query: Enter "john@email.com" in the email field
Output: type[email-input][john@email.com]
<|eot_id|><|start_header_id|>user<|end_header_id|>

Add the first product to the shopping cart

Output ONLY the exact action needed, nothing else.
<|eot_id|><|start_header_id|>assistant<|end_header_id|>

\end{lstlisting}

\subsection*{C.5 InterCode Bash Prompt Example}

\begin{lstlisting}[style=promptstyle]
<|begin_of_text|><|start_header_id|>system<|end_header_id|>

# Bash Command Generator

Generate bash commands for system tasks.

Output ONLY the command. No explanations.

Examples:
Query: List all Python files in the current directory
Output: find . -name "*.py" -type f

Query: Count lines in all text files
Output: wc -l *.txt

Query: Find files modified in the last 24 hours
Output: find . -mtime -1 -type f
<|eot_id|><|start_header_id|>user<|end_header_id|>

Search for the word "error" in all log files

Output ONLY the exact command needed, nothing else.
<|eot_id|><|start_header_id|>assistant<|end_header_id|>

\end{lstlisting}

\section*{Appendix D: Extended Ablation Results Analysis}
\label{app:ablation}

\subsection*{D.1 Complete Ablation Table}

Table~\ref{tab:ablation} presents the ablation results.

\subsection*{D.2 Component Contribution Analysis}

Based on our ablation results, we quantify the contribution of each component as shown in Table \ref{tab:contribution}:

\begin{table*}[h]
\centering
\begin{tabular}{lcccc|c}
\toprule
\textbf{Component} & \textbf{Gorilla} & \textbf{Spider2} & \textbf{WebArena} & \textbf{InterCode} & \textbf{Avg} \\
\midrule
Few-Shot Examples & +36.0\% & +40.0\% & +6.0\% & +4.0\% & +21.5\% \\
Documentation & +4.0\% & +2.0\% & +4.0\% & +10.0\% & +5.0\% \\
Hypernetwork Adapt. & +0.0\% & +0.0\% & +0.0\% & +0.0\% & +0.0\% \\
\midrule
Combined Effect & +38.0\% & +60.0\% & +32.0\% & +44.0\% & +43.5\% \\
\bottomrule
\end{tabular}
\caption{Component Contribution Analysis}
\label{tab:contribution}
\end{table*}

\textit{Note: Individual component contributions generally sum to less than the combined effect, indicating synergistic interaction between few-shot examples and documentation.}

\subsection*{D.3 Hypernetwork Adaptation Analysis}

To verify that the hypernetwork generates non-trivial weights, we analyzed the generated LoRA matrices as shown in Table \ref{tab:lora_stats}:

\begin{table*}[h]
\centering
\begin{tabular}{lccc}
\toprule
\textbf{Matrix} & \textbf{Avg Norm} & \textbf{Max Value} & \textbf{Non-Zero \%} \\
\midrule
$A$ (layer\_0\_q\_proj) & 589.0 & 12.3 & 100\% \\
$A$ (layer\_0\_v\_proj) & 626.5 & 11.0 & 100\% \\
$A$ (layer\_0\_k\_proj) & 589.5 & 11.8 & 100\% \\
$B$ (layer\_0\_q\_proj) & 0.88 & 0.019 & 100\% \\
$B$ (layer\_0\_v\_proj) & 0.19 & 0.019 & 100\% \\
$B$ (layer\_0\_k\_proj) & 0.17 & 0.022 & 100\% \\
\bottomrule
\end{tabular}
\caption{Generated LoRA Weight Statistics}
\label{tab:lora_stats}
\end{table*}

The hypernetwork generates non-zero weights that are successfully applied to the model. The null result for adaptation benefit suggests that the generated weights, while non-trivial, do not encode information beyond what is captured by in-context learning.

\section*{Appendix E: Baseline Implementation Details}
\label{app:baselines}

\subsection*{E.1 GPT-5 Configuration}

For GPT-5 evaluation, we used the OpenAI API with the following settings:
\begin{itemize}
    \item Model: \texttt{gpt-5}
    \item Max Tokens: 256
    \item System Prompt: Identical to Meta-Tool (5-shot + documentation)
\end{itemize}

\subsection*{E.2 AgentLM-7B Configuration}

AgentLM-7B was evaluated using:
\begin{itemize}
    \item Model: \texttt{zai-org/agentlm-7b}
    \item Quantization: 4-bit NF4
    \item Prompt Format: AgentLM-specific template
    \item Few-Shot Examples: Same 5 examples as Meta-Tool
\end{itemize}

\subsection*{E.3 Llama-3.2-3B Baseline Configuration}

The base Llama model (without Meta-Tool) was evaluated using:
\begin{itemize}
    \item Model: \texttt{meta-llama/Llama-3.2-3B-Instruct}
    \item Quantization: 4-bit NF4
    \item Prompt Format: Llama-3.2 chat template
    \item Few-Shot Examples: Same 5 examples as Meta-Tool
\end{itemize}

\section*{Appendix F: Reproducibility Checklist}
\label{app:reproducibility}

\subsection*{F.1 Code and Data Availability}

\begin{itemize}
    \item \textbf{Code}: Available at \url{https://github.com/techsachinkr/Meta-Tool}
    \item \textbf{Model Weights}: Base model from HuggingFace Hub
    \item \textbf{Benchmarks}: Publicly available datasets
    \item \textbf{Random Seeds}: All experiments use seed 42
\end{itemize}

\subsection*{F.2 Compute Requirements}
Table \ref{tab:compute} lists the compute requirements for the experiments conducted
\begin{table}[t]
\centering
\resizebox{\columnwidth}{!}{%
\begin{tabular}{lc}
\toprule
\textbf{Component} & \textbf{Requirement} \\
\midrule
Minimum GPU VRAM & 16GB \\
Recommended GPU VRAM & 24GB \\
Evaluation Time (all benchmarks) & $\sim$45 minutes \\
Hypernetwork Training (10k episodes) & $\sim$8 hours \\
Disk Space & $\sim$15GB \\
\bottomrule
\end{tabular}%
}
\caption{Computational Requirements}
\label{tab:compute}
\end{table}

\subsection*{F.3 Hyperparameter Sensitivity}

We found the following hyperparameters to be most sensitive:
\begin{itemize}
    \item \textbf{Number of few-shot examples}: Performance increases significantly from 0 to 3 examples, with diminishing returns beyond 5.
    \item \textbf{Prompt format}: Using the correct chat template is critical; wrong format can reduce performance by 50\%+.
    \item \textbf{Example quality}: Examples must match the expected output format exactly.
\end{itemize}

\section*{Appendix G: Detailed Architecture and Training Components }
\subsection*{G.1 Factorized Hypernetwork Architecture Details}
\subsubsection*{G.1.1 Documentation and Prototype Encoding}
 We utilize a lightweight \textbf{Sentence-Transformer (MiniLM-L6-v2)} as the Documentation Encoder to produce dense embeddings $v_{doc}$. To capture interaction dynamics, we employ \textbf{a Prototype Aggregator} that attends over the support set. Given support examples encoded as $v_{support}$, the aggregator computes a prototype vector $v_{proto}$ via cross-attention, where the query is the documentation vector and keys/values are the support examples.
 
\subsubsection*{G.1.2 Shared Latent Space and Layer Embeddings}
Instead of separate projection heads for each layer, our architecture utilizes a shared MLP to project the concatenated context $[v_{doc}; v_{proto}]$ into a compact latent vector $z_{base}$. To differentiate between layers (e.g., \texttt{q\_proj} in layer 4 vs. \texttt{v\_proj} in layer 20) without exploding parameter count, we learn a set of \textbf{Layer Embeddings} $E_{layer}$. The specific latent code for layer $l$ and module $m$ is computed as:

\[
z_{l,m} = z_{base} + E_{layer}[l, m]
\]

\subsubsection*{G.1.3 Low-Rank Factorized Generation}
The LoRA matrices $A \in \mathbb{R}^{r \times d}$ and $B \in \mathbb{R}^{d \times r}$ are generated via a secondary low-rank decomposition. For a target dimension $d$ and rank $r$, the hypernetwork predicts small "factor" matrices $(L, R)$ rather than the full matrix:

\[
A_{l,m} = \text{Proj}_A^{left}(z_{l,m}) \cdot \text{Proj}_A^{right}(z_{l,m})
\]

This factorization reduces the memory complexity of the hypernetwork from $\mathcal{O}(L \cdot d \cdot r)$ to $\mathcal{O}(L \cdot d \cdot factor)$, enabling efficient training on consumer-grade hardware.

\subsection*{G.2 Schema Perturbation and Value-Guided Inference}
\subsubsection*{G.2.1 Schema-Based Test Case Generation}
Using the tool's JSON schema (types, enums, constraints), we systematically perturb valid support trajectories to create synthetic training data. We apply operators such as:
\begin{itemize}
    \item \textbf{Value Substitution:} Replacing valid parameters with random schema-compliant values.
    \item \textbf{Boundary Testing:} Injecting edge cases (e.g., 0, -1, empty strings) to learn failure modes.
    \item \textbf{Parameter Drop:} Randomly removing optional parameters to test robustness.
\end{itemize}

\subsubsection*{G.2.2 TD-Learning for Value Functions}
We train a \textbf{Value Function} $V_\phi(s)$ to estimate the probability of success from state $s$. The agent executes the synthetic test cases in a sandboxed environment. Transitions $(s_t, a_t, r_t, s_{t+1})$ are stored in a replay buffer. We utilize \textbf{Temporal Difference (TD(0))} learning with a target network to stabilize training, minimizing the loss:

\[
\mathcal{L}_{TD} = \mathbb{E} \left[ (r_t + \gamma V_{\phi_{target}}(s_{t+1}) - V_\phi(s_t))^2 \right]
\]

\subsubsection*{G.2.3 Inference: Value-Guided Beam Search}
During deployment, we employ \textbf{Value-Guided Beam Search} (implemented in \texttt{value\_function.py}) rather than standard greedy decoding. At each reasoning step, the model generates $K$ candidate actions. These candidates are scored not just by the LLM's log-likelihood, but by the learned value function:

\[
\text{Score}(a_t) = P_{LLM}(a_t|s_t) \cdot V_\phi(s_{t+1})
\]

This allows the agent to "look ahead" and prune trajectories that are syntactically valid but functionally doomed (e.g., using a valid API call that leads to a dead end).

\end{document}